\documentclass[letterpaper]{article} % DO NOT CHANGE THIS
\usepackage{aaai24}  % DO NOT CHANGE THIS
\usepackage{times}  % DO NOT CHANGE THIS
\usepackage{helvet}  % DO NOT CHANGE THIS
\usepackage{courier}  % DO NOT CHANGE THIS
\usepackage[hyphens]{url}  % DO NOT CHANGE THIS
\usepackage{graphicx} % DO NOT CHANGE THIS
\usepackage{natbib}  % DO NOT CHANGE THIS AND DO NOT ADD ANY OPTIONS TO IT
\usepackage{caption} % DO NOT CHANGE THIS AND DO NOT ADD ANY OPTIONS TO IT
\usepackage{algorithm}

\usepackage{algorithmic}
\usepackage{epsfig}
\usepackage{graphicx}
\usepackage{amsmath}
\usepackage{amssymb}
\usepackage{subcaption}
\usepackage[font=small]{caption}

\usepackage{arydshln}

\usepackage{booktabs}
\usepackage{array}
\usepackage{multirow}
\usepackage{xcolor}
\usepackage{bm}
\usepackage{newfloat}
\usepackage{listings}
\DeclareCaptionStyle{ruled}{labelfont=normalfont,labelsep=colon,strut=off} % DO NOT CHANGE THIS
\floatstyle{ruled}
\newfloat{listing}{tb}{lst}{}
\floatname{listing}{Listing}
\title{Domain Generalizable Person Search Using Unreal Dataset}
\author{
    Minyoung Oh\equalcontrib,
    Duhyun Kim\equalcontrib,
    and Jae-Young Sim\thanks{Corresponding author.}\\
}
\affiliations{
    Graduate School of Artificial Intelligence, UNIST, Republic of Korea\\
    \{mmyy2513, ~duhyunkim, ~jysim\}@unist.ac.kr
}

\begin{document}

\maketitle
\begin{abstract}
Collecting and labeling real datasets to train the person search networks not only requires a lot of time and effort, but also accompanies privacy issues. 
The weakly-supervised and unsupervised domain adaptation methods have been proposed to alleviate the labeling burden for target datasets, however, their generalization capability is limited. 
We introduce a novel person search method based on the domain generalization framework, that uses an automatically labeled unreal dataset only for training but is applicable to arbitrary unseen real datasets. 
To alleviate the domain gaps when transferring the knowledge from the unreal source dataset to the real target datasets, we estimate the fidelity of person instances which is then used to train the end-to-end network adaptively. 
Moreover, we devise a domain-invariant feature learning scheme to 
encourage the network to suppress the domain-related features.
Experimental results demonstrate that the proposed method provides the competitive performance to existing person search methods even though it is applicable to arbitrary unseen datasets without any prior knowledge and re-training burdens. 
\end{abstract}

\section{Introduction}\label{sec:intro}
\begin{figure}[t]
    \centering
    \includegraphics[width=1.0\linewidth]{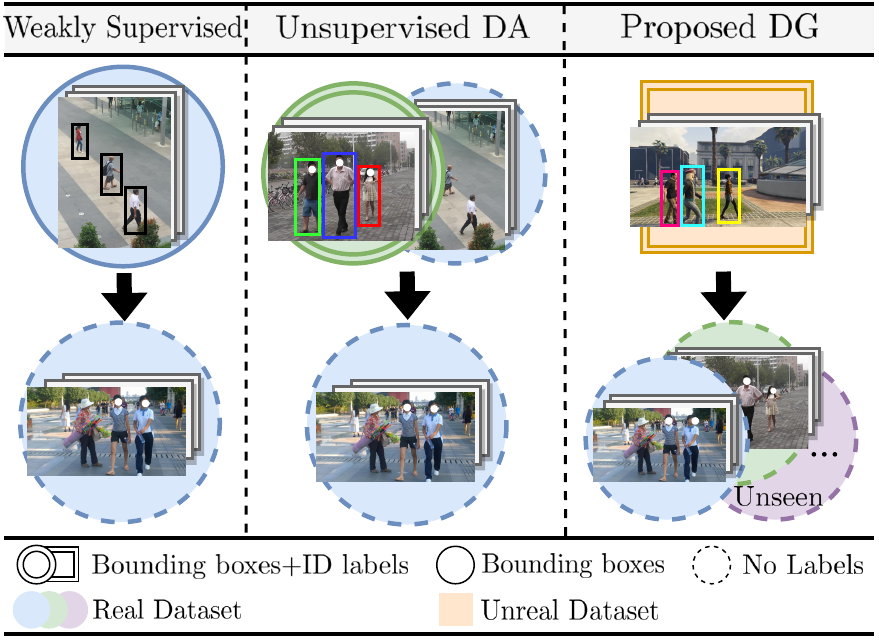}
    \caption{\normalsize{The proposed domain generalization concept compared to the weakly supervised and unsupervised domain adaptation methods. The upper and lower figures represent the training datasets and the test datasets, respectively.} }
\label{fig: figure1}
\end{figure}
Person search is a technique to detect the person instances from the scene images first, and then find a query person among the detected instances. Recently, it has been drawing a lot of attention in various computer vision applications such as surveillance and life logging. 
In general, large datasets of labeled scene images, captured under diverse environments, are required to train the person search networks. However, collecting such datasets is a time-consuming task, and furthermore, it usually requires a great deal of effort to obtain the ground truth labels by human annotation such as the bounding boxes and identities of persons. In addition, real datasets including personal information often suffer from the privacy issues.

To reduce the burden of data labeling, attempts have been made such as weakly supervised learning~\cite{bmvc, siamese, aaaiweakly} and unsupervised domain adaptation (DA)~\cite{uda}, whose concepts are compared in Figure~\ref{fig: figure1}. 
The weakly supervised methods assume that only the bounding box labels are given without the ID labels in the training dataset. 
On the other hand, the unsupervised DA method considers source and target datasets, respectively, where the source dataset has the labels of both the bounding boxes and identities, but the target dataset has no labels at all. 
It uses the labeled source dataset and the unlabeled target dataset together for training. 
However, both approaches are not fully generalizable to be directly applied to arbitrary unseen datasets without additional training burdens, since they still require partial labels and/or need to re-train the networks for a given target dataset.

\begin{figure}[]
	\centering
	\subfloat{
		\includegraphics[width=0.46\linewidth]{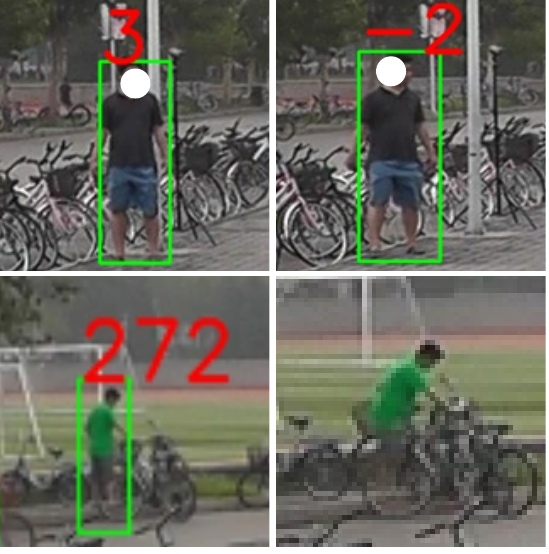}}\quad
	\subfloat{
		\includegraphics[width=0.46\linewidth]{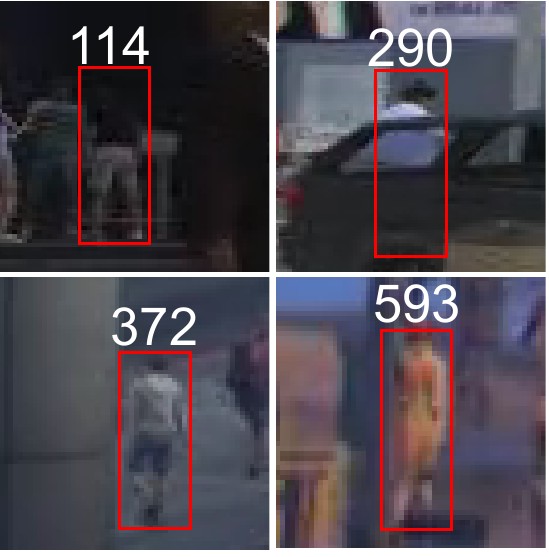}}
	\caption{\normalsize{The characteristics of the real PRW (left) and unreal JTA (right) datasets. The identity labels of persons are shown at the top of the bounding boxes.}
	}
	\label{fig: figure2}
\end{figure}

We propose a fully generalizable person search framework based on domain generalization (DG) from unreal dataset to arbitrary real datasets. 
In practice, we employ the unreal dataset of JTA (Joint Track Auto)~\cite{jta}, where the detailed labels were automatically annotated, as the only source dataset used for training. 
Then we test the trained network on arbitrary unseen target datasets captured in real environment. By using the unreal dataset, we are free from the time-consuming and labor-intensive labeling burdens as well as the privacy issues of real datasets. 
However, the knowledge transfer from an unreal dataset to the real datasets suffers from huge domain gaps that usually degrade the performance of person search. 
Specifically, we observe that manually annotated datasets often include incorrectly labeled and/or unlabeled instances, as shown in Figure~\ref{fig: figure2} (a). 
On the other hand, the automatically labeled unreal dataset always provides the correct labels even for some instances with degraded visibility due to severe occlusion, low contrast, or low resolution, as shown in Figure~\ref{fig: figure2} (b). 
To alleviate the domain gaps of annotation between the unreal and real datasets, we estimate the fidelity of each person instance using the deep features, which is used for fidelity adaptive training. 
Moreover, we regard each sequence in the unreal training dataset as each domain, and force the network to learn the domain-invariant features while disentangling the domain-specific information from the ID-specific features.

The main contributions of this paper are as follows. 
\begin{itemize}
    \item To the best of our knowledge, we first propose a novel framework of generalizable person search where only an unreal dataset is used for training, and arbitrary unlabeled real datasets can be tested at the inference phase.
    \item We develop the fidelity adaptive training and domain-invariant feature learning to alleviate the domain gaps between the unreal and real datasets improving the generalization capability.
    \item We show that the proposed method provides the competitive performance to the existing weakly-supervised and unsupervised DA methods, even though it is free from the re-training burdens and privacy issues.
\end{itemize}

\section{Related Work}
\subsubsection{Person Search}
The supervised methods of person search have been proposed that require labor-intensive labeling burdens. 
Xiao et al.~provided CUHK-SYSU~\cite{oim} dataset with the annotated ground truth labels of bounding boxes and identities. They proposed an end-to-end framework where the detection and re-identification networks are trained simultaneously. 
Zheng et al.~introduced PRW~\cite{prw} dataset with the annotated labels. They reflected the detection confidence to improve the re-identification accuracy. 
Chen et al.~decomposed the feature vector of each person instance into the norm and angle to overcome the conflict problem between the detection and re-identification tasks~\cite{nae}. 
Li and Miao performed the detection and re-identification progressively using the additional Faster-RCNN head~\cite{seqnet}. 
Han, Ko, and Sim adopted a part classifier network to prevent the overfitting and trained the network by weighting the detection confidence adaptively~\cite{agwf}. 
Yu et al.~tackled the occlusion problem by exchanging the tokens between the proposals based on the transformer~\cite{coat}.

To overcome the labeling burdens, the weakly supervised person search methods have been studied that assume only the bounding boxes are labeled without ID labels. 
Han, Ko, and Sim devised a context-aware clustering method using the uniqueness property that multiple persons in a certain scene image do not have the same ID, and the co-appearance property where the neighboring persons tend to appear simultaneously~\cite{bmvc}. 
Han et al.~trained the network to yield more reliable results of re-identification by using both features from the scene-level proposals and the cropped bounding boxes~\cite{siamese}. 
Yan et al.~used the context information to enhance the clustering accuracy~\cite{aaaiweakly}. 
On the other hand, the unsupervised person search method was also introduced based on DA~\cite{uda}, which uses the labeled source dataset and unlabeled target dataset together for training. 
However, the weakly supervised methods still need partial labels of target datasets, and the unsupervised method should re-train the network whenever a target dataset is newly given. 

\subsubsection{Domain Generalization}
The DG techniques aim to design robust networks when tested on any unseen dataset while using the limited training datasets. 
There have been three main ways to improve the generalization capability of image classification and segmentation: data augmentation~\cite{dg_cvpr20,dg_iccv21,dg_nips18}, meta-learning~\cite{dg_aaai18,dg_neurips18,dg_cvpr21}, and representation learning~\cite{segu2022batch,Motiian_2017_ICCV,fan2021adversarially}. 

Recently, the DG techniques have been adopted for person re-identification tasks. 
Jin et al.~normalized the style variations across the different domains and restored the lost ID-related information caused by the instance normalization~\cite{snr}.
Choi et al.~adopted the batch-instance normalization layers trained with the meta-learning strategy to avoid the overfitting to the source domain~\cite{metabin}. 
Liu et al.~defined a hybrid domain composed of the datasets from multiple domains, and trained the dataset in the hybrid domain with an ensemble of other batch normalization parameters to encourage the generalization capability~\cite{reid_aaai22}. 
However, these methods were devised for re-identification and cannot be directly applied to the person search task combined with the addition person detection.

\section{Unreal Dataset}

\begin{figure}[]
	\centering
	\begin{subfigure}{0.32\linewidth}
		\includegraphics[width=0.99\linewidth]{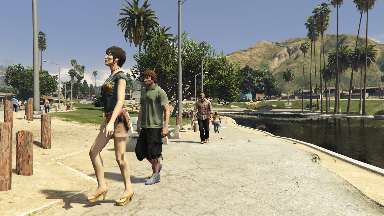}
	\end{subfigure}
	\begin{subfigure}{0.32\linewidth}
		\includegraphics[width=0.99\linewidth]{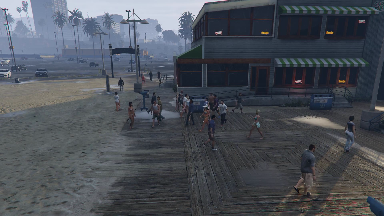}
	\end{subfigure}
	\begin{subfigure}{0.32\linewidth}
		\includegraphics[width=0.99\linewidth]{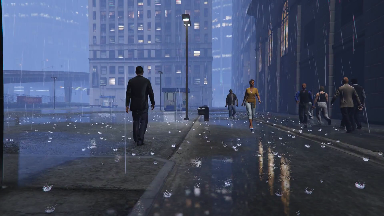}
	\end{subfigure}
	\begin{subfigure}{0.32\linewidth}
		\includegraphics[width=0.99\linewidth]{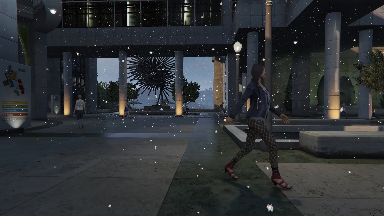}
	\end{subfigure}
	\begin{subfigure}{0.32\linewidth}
		\includegraphics[width=0.99\linewidth]{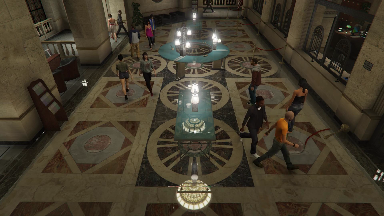}
	\end{subfigure}
	\begin{subfigure}{0.32\linewidth}
		\includegraphics[width=0.99\linewidth]{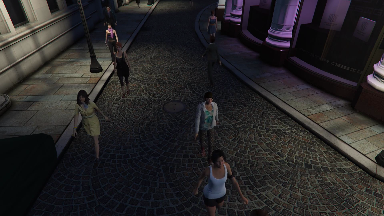}
	\end{subfigure}
	\caption{\normalsize{Images from the unreal JTA dataset.}}
	\label{fig: figure3}
\end{figure} 
We used the unreal dataset of JTA~\cite{jta} obtained from the photo-realistic video game Grand Theft Auto V, where the details for each person instance are automatically annotated such as the bounding boxes, identities, and keypoints. JTA provides about 450,000 images extracted from 512 video sequences with diverse characteristics, such as the background, viewpoint, and weather condition, as shown in Figure~\ref{fig: figure3}.
We constructed the JTA* dataset based on JTA for the purpose of person search by taking 256 sequences in the training category of JTA,  which are then divided into 226 sequences for training and 30 sequences for test, respectively. We selected every tenth image from the training sequences and every sixth image from the test sequences, respectively. 
JTA* is expected to serve as a more reliable training dataset, since many different identities can be used as the negative samples to improve the performance from the perspective of contrastive learning~\cite{oim, simclr}.
Moreover, JTA* has no incorrectly labeled or unlabeled instances at all with the help of automatic annotation. 
However, the unreal dataset does not completely capture the styles of real-world scenes in general, which makes it hard to transfer the knowledge learned from the unreal dataset to the real datasets. For example, the instances with severely degraded visibility tend to be undetected as persons in real datasets. Accordingly, using all the instances in unreal dataset for training may degrade the performance of person search when tested in real datasets. Therefore, we only used the person instances, where the numbers of occluded keypoints are less than 13, to exclude severely occluded instances from training.
Table \ref{tab: table1} shows the specifications of JTA* that exhibits relatively larger numbers of person identities and instances compared to the other existing real datasets of CUHK-SYSU and PRW. 

\begin{table}[]
    \small{
    \begin{tabular}{ccccc}
    \toprule
          &  & { JTA*} & { CUHK-SYSU} & { PRW}\\
         \midrule\midrule[.1em]
     \multirow{2}{1.4cm}[-0.4ex]{ \#Images} 
        & { Train} & { 10,049} & { 11,206} & { 5,134} \\
     \cmidrule{2-5}
        & { Test}     & { 4,426}  & { 6,978}  & { 6,112} \\
    \midrule
     \multirow{2}{1.4cm}[-0.4ex]{ \#Persons} 
        & { Train} & {175,035} & { 55,272} & { 16,243} \\
     \cmidrule{2-5}
        & { Test} & { 74,382} & { 40,871} & { 25,062} \\ 
    \midrule
     \multirow{2}{1.4cm}[-0.4ex]{ \#IDs} 
        & { Train} & {10,912} & { 5,532} & { 482} \\
     \cmidrule{2-5}
        & { Test} & { 1,480} & { 2,900} & { 450} \\
    % \midrule
    \bottomrule
    \end{tabular}}
    \caption{\normalsize{The specifications of datasets.}}
    \label{tab: table1}
\end{table}

\begin{figure*}[t]
	\centering
	\includegraphics[width=0.94\linewidth]{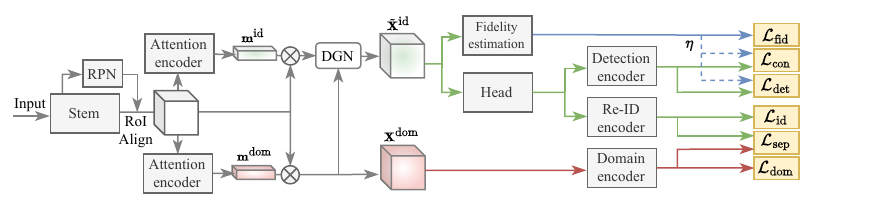}
	\caption{\normalsize{The overall framework of the proposed method. At the training phase, the ID-specific and domain-specific features are extracted by using the attention encoders where the ID-specific features are used to estimate the fidelity of person instance. The estimated fidelity is then used to adaptively compute $\mathcal{L}_\mathrm{det}$ and $\mathcal{L}_\mathrm{con}$ in the head network. The domain-specific features are used to calculate $\mathcal{L}_\mathrm{dom}$ and $\mathcal{L}_\mathrm{sep}$. At the inference phase, only the ID-specific features are used. The dashed lines indicate the stop-gradient operation.}} 
	\label{fig: figure4}
\end{figure*}

\section{Method}\label{sec:method}
We train the unreal dataset JTA* as the only source dataset, and test the trained network on arbitrary real target datasets based on the DG framework for person search.
To alleviate the domain gaps between the unreal and real datasets, we first estimate the fidelity of person instance in JTA* by extracting the deep features. Then we use the estimated fidelity to adaptively train the network suppressing the influence of degraded person instances which are difficult to be identified.
Furthermore, we also improve the generalization capability of network by disentangling the domain and ID-specific features to reduce the dependency on the domain information.
Figure~\ref{fig: figure4} shows the overall framework of the proposed method. 

\subsection{Fidelity Adaptive Training}
When all the instances with degraded visibility in the unreal dataset are used for training, the network tends to overfit to the source dataset and thus yields low performance on real target datasets. 
We may remove such degraded instances from the training dataset by using the automatically annotated information, e.g., the keypoints with occlusion information and the size of bounding box. However, it is not trivial to set a criterion for the fidelity of instance in terms of the performance of person search. Moreover, some of the degraded instances in the training dataset may help to improve the robustness of the network to identify the challenging person instances in real datasets. In order to strike a balance between suppressing the effect of degraded instances and improving the robustness of network, we estimate the fidelity of person instances which is then used to train the network adaptively. 

\subsubsection{Fidelity Estimation.}
\begin{figure}[]
	\centering
		\includegraphics[width=0.97\linewidth]{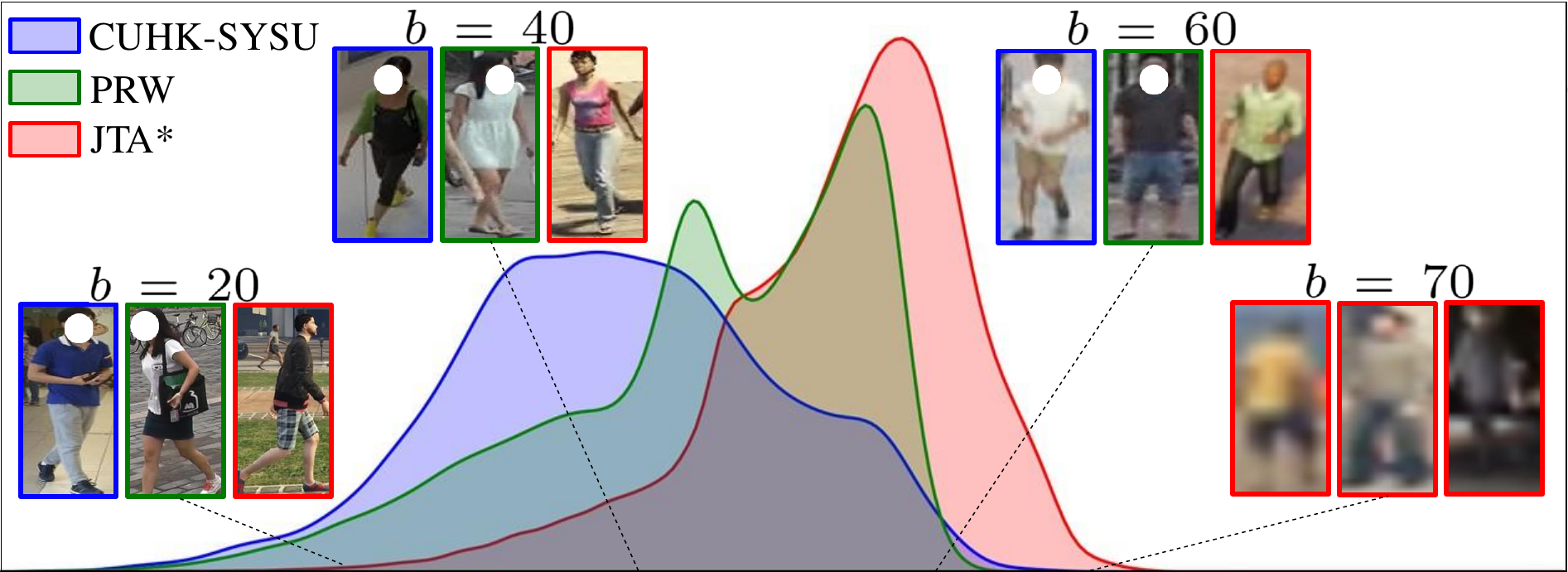}
	\caption{\normalsize{The BRISQUE score distributions for the cropped images of person instances.}}
	\label{fig: figure5}
\end{figure}

We basically estimate the fidelity as the visibility or quality of the image. To this end, we use the BRISQUE~\cite{brisque} which measures the naturalness of image. The high values of BRISQUE score represent severely distorted or noisy images. Figure~\ref{fig: figure5} compares the distributions of BRISQUE scores computed over the bounding box images of person instances among the three datasets of CUHK-SYSU, PRW, and JTA*. Whereas most of the person instances in the real datasets of CUHK-SYSU and PRW have lower BRISQUE scores than 60, many instances in the unreal dataset of JTA* exhibit relatively higher scores. For example, we see blurred and/or low contrast images at the scores around 70.

We train the fidelity estimation network, composed of the four convolutional layers and a fully-connected layer, by minimizing the fidelity estimation loss $\mathcal{L}_\mathrm{fid}$.
\begin{equation}
\mathcal{L}_\mathrm{fid} = \frac{1}{\left| \Omega_\mathrm{p} \right|}\sum_{i \in \Omega_\mathrm{p}} {\left\{ {\eta_i} - \mathrm{exp}(-b_i/\tau_\mathrm{fid}) \right\}}^2,
\label{eq: equation1}
\end{equation}
where ${\eta_i}$ denotes the estimated fidelity of the $i$-th person instance, $b_i$ is the BRISQUE score measured on the $i$-th person instance, $\tau_\mathrm{fid}$ is a hyperparameter, and $\Omega_\mathrm{p}$ denotes the index set of the predicted person instances among all the proposals in a batch. 
Note that we do not determine the fidelity directly from the BRISQUE score, but we estimate the fidelity values by extracting the deep features. It means that even the person instances with similar BRISQUE scores may be assigned largely different fidelity values according to their actual appearance or visibility. Therefore, the proposed network flexibly learns the relationship between the level of naturalness of image and the actual fidelity.

\subsubsection{Fidelity Weighted Detection Loss.}
The degraded person instances with low fidelity values often make the detection network confuse the true positive instances with the false positive ones at test time. To alleviate such effect, we reflect the fidelity to adaptively compute the multi-task detection losses of Faster R-CNN~\cite{fasterrcnn}. Specifically, we modify the detection loss $\mathcal{L}_\mathrm{det}$ such that the classification loss, associated with the foreground instances, is weighted by the estimated fidelity which is fixed by the stop-gradient operation. Note that when a person instance yields a low fidelity value, the associated classification loss is decreased during the training, and the contribution of this instance is suppressed accordingly.

\subsubsection{Fidelity Guided Confidence Loss.}
We adopt the adaptive gradient weighting function~\cite{agwf} to reflect the contribution of each instance to the training adaptively. However, when we train the network using all the labeled instances in the unreal dataset regardless of their visibility, the network may assign relatively high detection confidence to the instances with degraded visibility. In such a case, it becomes difficult to fully enjoy the benefits of the adaptive gradient weighting function. Therefore, we utilize the estimated fidelity to supervise the confidence scores as well, to avoid severely degraded instances from having abnormally high confidence scores. Specifically, we design the fidelity-guided confidence loss
\begin{equation}
	\mathcal{L}_{\mathrm{con}} = \frac{1}{\left|\Omega_\mathrm{p}\right|} \sum_{i \in \Omega_\mathrm{p}} (\alpha_i - \bar{\eta}_i)^2,
	\label{eq: equation2}
\end{equation}
where $\alpha_i$ is the confidence probability of the $i$-th person instance, and $\bar{\eta}_i$ denotes the fixed fidelity value which is not updated during the gradient back-propagation by using the stop-gradient operation.

\subsubsection{Fidelity Weighted Feature Update.}
At each iteration, the ID-specific feature of a new instance is used to incrementally update the ID look-up table (ILT). However, the degraded instances usually exhibit not only the ID-specific features but also a considerable amount of the ID-unrelated features. Therefore, the ILT may not represent correct person identities when updated by using the degraded instances directly. To deal with this problem, we also utilize the estimated fidelity of person instances to update the feature vector in ILT such that 
\begin{equation}
	\bm{f}_{k_i}^{\rm id}\longleftarrow w_{\rm id}\bm{f}_{k_i}^{\rm id} + \bar{\eta}_i (1-w_{\rm id}) \bm{x}_i^{\mathrm{id}},
	\label{eq: equation3}
\end{equation}
where $\bm{f}_{k}^{\rm id}$ is the 256-dimensional feature vector corresponding to the $k$-th person ID in ILT, $\bm{x}_i^{\mathrm{id}}$ denotes the ID-specific feature of the $i$-th person instance, $k_i$ is the ground truth ID of the $i$-th person instance, and $w_{\rm id}$ is a momentum parameter.
We normalize the updated feature vector to have the length of 1. By using the fidelity, we can suppress the impact of the features obtained from the degraded person instances to update the ILT. 

We also use the re-identification loss $\mathcal{L}_{\mathrm{id}}$~\cite{oim} applying the adaptive gradient weighting function to learn the ID-discriminative features.
\begin{equation}	
	\mathcal{L}_{\mathrm{id}}= -\frac{1}{\left|\Omega_\mathrm{p}\right|} \sum_{i \in \Omega_\mathrm{p}} \log\frac{ \exp(\alpha_i \langle\bm{f}_{k_i}^{\rm id}, \bm{x}_i^{\mathrm{id}}\rangle / \tau_\mathrm{id})} {\sum_{j=1}^{L} \exp(\alpha_i\langle\bm{f}_{j}^{\rm id}, \bm{x}_i^{\mathrm{id}}\rangle / \tau_\mathrm{id})},
	\label{eq: equation4}
\end{equation}
where $\langle \cdot, \cdot\rangle$ is the inner-product operation, $L$ is the size of ILT, and $\tau_\mathrm{id}$ is a temperature parameter. Note that we do not employ the unlabeled identities to compute the re-identification loss in~\eqref{eq: equation4}, since all the person instances have the ground truth identities in the unreal dataset.

\subsection{Domain Invariant Feature Learning}
To overcome the domain gap when transferring the knowledge from the unreal dataset to the real datasets, we attempt to learn the domain-invariant and ID-specific features, respectively. We regard each sequence in the unreal dataset as a unique domain assuming that it represents different characteristics such as the background, viewpoint, and weather condition.
We employ the attention encoder network composed of the global average pooling and the convolutional layers to learn a channel attention vector. As shown in Figure~\ref{fig: figure4}, two attention vectors of $\mathbf{m}_{i}^\mathrm{id} \in \mathbb{R}^{c}$ and $\mathbf{m}_{i}^\mathrm{dom} \in \mathbb{R}^{c}$, where $c$ is the number of channels, are independently extracted from the $i$-th person instance. 

\subsubsection{Domain-Guided Feature Normalization.}
We multiply each element in $\mathbf{m}_{i}^\mathrm{id}$ and $\mathbf{m}_{i}^\mathrm{dom}$ with each feature map at the corresponding channel to extract the ID-specific feature map, $\mathbf{X}_{i}^\mathrm{id} \in \mathbb{R}^{c \times h \times w}$, and the domain-specific feature map, $\mathbf{X}_{i}^\mathrm{dom} \in \mathbb{R}^{c \times h \times w}$, respectively, where $h$ and $w$ indicate the height and width of the feature map.
To improve the domain-agnostic ID discriminative capability, we additionally normalize $\mathbf{X}_{i}^\mathrm{id}$ by using the statistics of $\mathbf{X}_{i}^\mathrm{dom}$ such that 
\begin{equation}
   \tilde{\mathbf{X}}_{i}^\mathrm{id} = \frac{\mathbf{X}_{i}^\mathrm{id} - \mu (\mathbf{X}_{i}^\mathrm{dom})}{\sigma(\mathbf{X}_{i}^\mathrm{dom})},
   \label{eq: equation5}
\end{equation}
where $\tilde{\mathbf{X}}_{i}^\mathrm{id}$ is the result of the domain-guided normalization (DGN), and $\mu$ and $\sigma$ denote the operations to calculate the mean and standard deviation at each channel of feature map, respectively.
Whereas $\tilde{\mathbf{X}}_{i}^\mathrm{id}$ is mapped into $\bm{x}_{i}^\mathrm{id}$ by the head network both at the training and test phases together, $\mathbf{X}_{i}^\mathrm{dom}$ is fed into the domain encoder network at the training phase only, yielding a domain-specific feature vector $\bm{x}_{i}^{\mathrm{dom}}$.

\subsubsection{Domain Separation Loss.}
Note that, when both $\mathbf{m}_{i}^\mathrm{id}$ and $\mathbf{m}_{i}^\mathrm{dom}$ become identical to each other, the DGN operation becomes equivalent to the instance normalization~\cite{instancenorm}. To exploit the benefit of DGN by learning the distinct features from each other, we suggest a domain separation loss given by
\begin{equation}
    \!\!\!\!\mathcal{L}_{\mathrm{sep}} = \mathrm{exp}\left(-\mathrm{MMD}^2(\{\bm{x}_{i}^{\mathrm{dom}}\},\{\bm{x}_{i}^{\mathrm{id}}\})\right)\text{, for }\forall i \in \Omega_\mathrm{p},\!\!
    \label{eq: equation6}
\end{equation}
where $\mathrm{MMD}(A,B)$ means the mean maximum discrepancy between two sets of $A$ and $B$~\cite{mmd}. By maximizing the difference of distribution between the sets of $\bm{x}_{i}^{\mathrm{dom}}$ and $\bm{x}_{i}^{\mathrm{id}}$, we force them construct unique distributions with respect to ID and domain, respectively.

\subsubsection{Domain Feature Update.}
To extract a representative $\bm{x}_{i}^{\mathrm{dom}}$ for its own domain, we first build a new domain look-up table (DLT) and update the DLT such that
\begin{equation}
	\bm{f}_{s_i}^{\rm dom} \longleftarrow w_{\rm dom}\bm{f}_{s_i}^{\rm dom} + (1-w_{\rm dom}) \bm{x}_{i}^{\mathrm{dom}},
	\label{eq: equation7}
\end{equation}
where $\bm{f}_{s}^{\rm dom}$ is the feature vector corresponding to the $s$-th element in DLT that represents the domain characteristics of the $s$-th sequence in the unreal training dataset, $s_i$ is the ground truth sequence label where $\bm{x}_{i}^{\mathrm{dom}}$ belongs to, and $w_\mathrm{dom}$ is a momentum parameter.
The updated feature vector $\bm{f}_{s}^{\rm dom}$ is also normalized to have the length of 1. 
Based on the DLT, we introduce the domain loss $\mathcal{L}_{\mathrm{dom}}$ as follows.
\begin{equation}
	\!\!\!\!\! \mathcal{L}_{\mathrm{dom}} \!\!=\! -\frac{1}{\left|\Omega_\mathrm{p}\right|} \!\! \sum_{i \in \Omega_\mathrm{p}} \! \log\frac{ \exp( \langle \bm{f}_{s_i}^{\rm dom}, \bm{x}^{\mathrm{dom}}_{i} \rangle/ \tau_\mathrm{dom})} {\sum_{j=1}^{D} \exp(\langle\bm{f}_{j}^{\rm dom}, \bm{x}_{i}^{\mathrm{dom}} \rangle / \tau_\mathrm{dom})},\!\!
	\label{eq: equation8}
\end{equation}
where ${D}$ is the size of DLT, and $\tau_\mathrm{dom}$ is a temperature parameter. By maximizing the cosine similarity of $\bm{x}^{\mathrm{dom}}_{i}$ to $\bm{f}_{s_i}^{\rm dom}$ while minimizing that to the others, $\mathcal{L}_{\mathrm{dom}}$ enhances the domain-specific representation of $\bm{x}^{\mathrm{dom}}_{i}$. 

\section{Experimental Results}
\subsection{Experimental Setup}
\subsubsection{Datasets.}
We used the unreal dataset of JTA* only for training, and used the real datasets of CUHK-SYSU~\cite{oim} and PRW~\cite{prw} for testing. CUHK-SYSU consists of various scene images captured with a moving camera and the frames selected from the movies and TV shows. PRW includes the images captured by 6 fixed cameras with different locations and viewing directions. The dataset specifications are summarized in Table~\ref{tab: table1}.

\subsubsection{Evaluation Measures.}
We used the Precision and Recall to evaluate the detection performance, and used the mean Average Precision (mAP) and Top-$1$ scores for re-identification performance. Only the proposals with larger than 0.5 IoU to the ground truth bounding boxes are used to evaluate the Top-$1$ scores.

\subsubsection{Implementation Details.}
We adopted the end-to-end person search network~\cite{nae} with the adaptive gradient weighting function~\cite{agwf} as a baseline, where the detection and re-identification networks are trained simultaneously.
We used PyTorch for all experiments with a single NVIDIA RTX-3090 GPU.
We used the ImageNet pre-trained ResNet50 as our backbone network for a fair comparison.
We set the batch size to 4 and used the SGD optimizer with a momentum of 0.9. The warm-up learning rate scheduler linearly increases the learning rate from 0 to 0.003 during the first epoch, and the learning rate decays by multiplying 0.1 every third epoch. We empirically set the weights of losses to 10 and 0.1 for $\mathcal{L}_{\mathrm{fid}}$ and $\mathcal{L}_{\mathrm{dom}}$, respectively, and 1 otherwise. $\tau_\mathrm{fid}=200$ in~\eqref{eq: equation1}, $\tau_\mathrm{id}=1/30$ in~\eqref{eq: equation4}, $\tau_\mathrm{dom}=1$ in~\eqref{eq: equation8}, $w_\mathrm{id}=2/3$ in~\eqref{eq: equation3}, and $w_\mathrm{dom}=2/3$ in~\eqref{eq: equation7}. During the training phase, we applied the Resize and HorizontalFlip transformations with the probability of 0.5.

\begin{table}[t]
	\centering
        \small{
	\begin{tabular} {@{\hspace{0mm}}l | @{\hspace{1mm}}c @{\hspace{2mm}}c @{\hspace{1mm}}| @{\hspace{1mm}}c @{\hspace{2mm}}c@{\hspace{1mm}}}
		\toprule
		\multicolumn{1}{c|@{\hspace{1mm}}}{\multirow{2}{*}{Method}} & \multicolumn{2}{c|@{\hspace{1mm}}}{CUHK}  & \multicolumn{2}{c}{PRW} \\  \cline{2-5}
        \multicolumn{1}{c|@{\hspace{1mm}}}{}& mAP     & Top-1   & mAP   & Top-1       \\ \midrule\midrule[.1em]
		{OIM  \cite{oim}}                  & 75.5    & 78.7    & 21.3  & 49.9  \\ 
		{HOIM \cite{hoim}}                 & 89.7    & 90.8    & 39.8  & 80.4  \\ 
        {NAE  \cite{nae}}                 & 92.1    & 92.9    & 44.0  & 81.1  \\ 
        {OIMNet++ \cite{oimnet++}}       & 93.1    & 93.9    & 46.8  & 83.9  \\ 
		{SeqNet  \cite{seqnet}}           & 94.8    & 95.7    & 47.6  & 87.6  \\ 
        {PSTR   \cite{pstr}}                & 93.5    & 95.0    & 49.5  & 87.8  \\ 
        {DMRNet++ \cite{dmrnet++}}          & 94.4   & 95.5     & 51.0  & 86.8  \\ 
		{COAT   \cite{coat}}               & 94.2    & 94.7    & 53.3  & 87.4  \\ 
		{AGWF  \cite{agwf}}     			   & 93.3    & 94.2    & 53.3  & 87.7  \\ 
        \midrule
		{CGPS  \cite{aaaiweakly}}          & 80.0    & 82.3    & 16.2  & 87.6  \\ 
		{R-SiamNet \cite{siamese}}         & 86.0    & 87.1    & 21.4  & 75.2  \\ 
		{CUCPS \cite{bmvc}}                & 81.1    & 83.2    & 41.7  & 86.0  \\ 
        \midrule
		{Unsupervised DA\cite{uda}}                  & 77.6    & 79.6    & 34.7  & 80.6  
        \\ \midrule
		{Proposed DG}                              & 76.1    & 78.4    & 25.5  & 79.4  \\ \bottomrule
	\end{tabular}}
	\caption{\normalsize{Comparison of the quantitative performance. The supervised and weakly-supervised methods are grouped in the first and second categories, respectively.}}
 
\label{tab: table2}
\end{table}

\begin{table}[t]
	\centering
        \small{
	\begin{tabular}{@{\hspace{0mm}}l@{\hspace{1mm}}|@{\hspace{1mm}}c@{\hspace{1mm}}c@{\hspace{1mm}}c@{\hspace{1mm}}c@{\hspace{1mm}}|@{\hspace{1mm}}c@{\hspace{1mm}}c@{\hspace{1mm}}c@{\hspace{1mm}}c@{\hspace{0mm}}}
		\toprule
		\multicolumn{1}{c@{\hspace{1mm}}|@{\hspace{1mm}}}{\multirow{2}{*}{Method}} & \multicolumn{4}{c|@{\hspace{1mm}}}{CUHK-SYSU}  & \multicolumn{4}{c}{PRW} \\  \cline{2-9}
		     & mAP     & Top-1   & AP& Recall& mAP   & Top-1 & AP& Recall      \\ \midrule\midrule[.1em]
		HOIM                 & 38.5 & 42.5& 57.1& 81.4& 12.2 & 37.8  & 71.4&92.4   \\ 
		NAE                   & 40.8 & 44.9 & 57.1& 69.2& 14.1 & 42.1 & 65.6&80.0    \\
		SeqNet             & 62.3 & 65.1 & 56.3& 64.5& 19.2 & 74.7 & 77.2&88.3   \\ 
        OIMNet++         & 66.3 & 69.0 & 60.4& 69.7& 19.8 & 74.0 & 74.8&84.7 \\
		COAT                 & 61.4 & 64.7 & 57.0& 60.2& 22.6 & 76.9   &81.2&87.9   \\
		Proposed                         & \bf{76.1} & \bf{78.4} &\textbf{72.3}&\textbf{87.3}& \bf{25.5} & \bf{79.4}& \bf{84.8}& \bf{96.0}\\ 
        \bottomrule
	\end{tabular}}
	\caption{\normalsize{Comparison of the DG performance. All the methods were trained by using the JTA* dataset only.}}
\label{tab: table3}
\end{table}

\begin{table}[t]
	\centering
	\small{
		\begin{tabular}{@{\hspace{0mm}}l@{\hspace{1mm}}| @{\hspace{1mm}}c @{\hspace{1mm}}c @{\hspace{1mm}}c @{\hspace{1mm}}c@{\hspace{1mm}}|@{\hspace{1mm}}c@{\hspace{1mm}}c@{\hspace{1mm}}c@{\hspace{1mm}}c@{\hspace{0mm}}}
			\toprule
			\multicolumn{1}{@{\hspace{0mm}}c@{\hspace{1mm}}|@{\hspace{1mm}}}{\multirow{2}{*}{Method}} & \multicolumn{4}{c|@{\hspace{1mm}}}{CUHK-SYSU}  & \multicolumn{4}{c@{\hspace{0mm}}}{PRW} \\  \cline{2-9}
			& mAP     & Top-1   & AP& Recall& mAP   & Top-1 & AP& Recall      \\ \midrule\midrule[.1em]
			Baseline             & 66.7 & 71.0 & 64.6& 78.5 & 20.9 & 76.0 & 77.8&92.3   \\ 
			w/ FAT             & 75.8 & \textbf{78.5} &68.3&87.3& 21.5 & 77.9 &82.3&95.9   \\ 
			w/ DIL         & 69.3 & 72.7 &66.8&80.2& 24.8 & \textbf{79.8} &79.8&92.7 \\
			Proposed       & \bf{76.1} & {78.4} &\textbf{72.3}&\textbf{87.3}& \bf{25.5} & {79.4}&\textbf{84.8}&\textbf{96.0}\\ 
			\bottomrule
	\end{tabular}}
	\caption{\normalsize{Ablation study of the proposed method.}}
	\label{tab: table4}
\end{table}

\begin{table}[t]
	\centering
	\small{
	\begin{tabular}{c|c|cc|cc}
		\toprule
		{\multirow{2}{*}{Method}} & \multirow{2}{*}{Resize}  & \multicolumn{2}{c|}{CUHK-SYSU}  & \multicolumn{2}{c}{PRW} \\  \cline{3-6}
		& & {mAP} & {Top-1} & {mAP} & {Top-1} \\
		\midrule\midrule[.1em]
		{\multirow{2}{*}{Baseline}} &             & {62.6} & {66.7} & \bf{21.0} & \bf{76.9} \\
		{} & \checkmark  & \bf{66.7} & \bf{71.0} & {20.9} & {76.0} \\ \midrule
		{\multirow{2}{*}{Proposed}} &             & {72.3} & {74.9} &{25.0} & \bf{80.8} \\ 
		{} & \checkmark  & \bf{76.1} & \bf{78.4} &\bf{25.5} & {79.4} \\
		\bottomrule
	\end{tabular}}
	\caption{\normalsize{Effect of the resize transformation.}}
	\label{tab: table5}
\end{table}

\begin{table}[t!]
	\centering
	\small{
		\begin{tabular}{ccc|cc|cc}
			\toprule
			\multirow{2}{*}{FWDL}& \multirow{2}{*}{FGCL} & \multirow{2}{*}{FWFU} & \multicolumn{2}{c|}{CUHK-SYSU} & \multicolumn{2}{c}{PRW} \\ \cline{4-7}
			& & & {mAP} & {Top-1} & {mAP} & {Top-1}\\
			\midrule\midrule[.1em]
			{ } & { } & { }                                  & {66.7} & {71.0}        & {20.9} & {76.0} \\ 
			{\checkmark } & { } & { }                        & {75.5} & {78.0}        & {21.4} & {77.9} \\
			{} & {\checkmark } & { }                         & {71.2} & {74.4}        & {21.1} & {77.5} \\
			{} & {} & {\checkmark  }                         & {71.7} & {75.1}        & {21.3} & {77.4} \\
			{\checkmark} & {\checkmark} & {\checkmark }      & \bf{75.8} & \bf{78.5}  & \bf{21.5} & \bf{77.9} \\ \midrule
			{$\dagger$\checkmark } & {$\dagger$\checkmark } & {$\dagger$\checkmark }    & {74.4} & {77.5}        & {21.3} & {77.8} \\ 
			\bottomrule
	\end{tabular}}
	\caption{\normalsize{Ablation study of fidelity-weighted detection loss (FWDL), fidelity-guided confidence loss (FGCL), and fidelity-weighted feature update (FWFU). $\dagger$\checkmark~indicates that we use the pre-defined ground-truth fidelity instead of the learned fidelity.}} % instead of the learned fidelity
	\label{tab: table6}
\end{table}

\subsection{Performance Comparison}
Note that the proposed framework of domain generalizable person search is first introduced in this paper, and there is no existing method fairly comparable to the proposed one. Instead, we compared the quantitative performance of the proposed method and the existing person search methods with different experimental settings including the supervised, weakly-supervised, and unsupervised DA methods, as shown in Table~\ref{tab: table2}. Whereas all the compared existing methods use the target test dataset for training in any way, the proposed method does not access test datasets at all during training. Nevertheless, the proposed method provides the comparable performance to the existing methods and even surpasses several supervised and weakly-supervised methods on both target datasets. 

In addition, we also compared the DG performance of several supervised methods and the proposed one in Table~\ref{tab: table3}. We trained all the compared networks by using the JTA* dataset only. We see that the proposed method achieves a much higher performance of DG compared with the existing methods. Consequently, the experimental results demonstrate that the proposed method is a promising technique for person search which is completely free from the burden of time-consuming and labor-intensive labeling as well as the privacy issues.

\subsection{Ablation Study}
We validated the effectiveness of the proposed fidelity adaptive training (FAT) and domain-invariant feature learning (DIL), respectively. Table~\ref{tab: table4} demonstrates that each of FAT and DIL improves not only the re-identification performance but also the detection performance.

\subsubsection{Effect of Resize Transformation.}
There are huge differences in the size and aspect ratio of image between CUHK-SYSU and JTA* datasets. While the image size and aspect ratio of JTA* are fixed to 1920 $\times$ 1080 and 1.78, respectively, CUHK-SYSU dataset has very diverse image sizes and aspect ratios. This discrepancy between the source and target datasets can be another source of domain gap. Therefore, when training the JTA* dataset, we applied the resize transformation with 0.5 probability to prevent the model from overfitting to the source dataset with the fixed image size. Table~\ref{tab: table5} shows that the resize transformation keeps the performance from degradation caused by the size difference in CUHK-SYSU dataset. On the other hand, most of the images in PRW dataset have the same size to that of JTA*, and thus the resize transformation does not yield a significant performance gain when applied on PRW dataset.

\subsubsection{Effect of Fidelity Adaptive Training.}
Table~\ref{tab: table6} shows the detailed results of ablation study for the three schemes of FAT: fidelity-weighted detection loss (FWDL), fidelity-guided confidence loss (FGCL), and fidelity-weighted feature update (FWFU), where we see that each scheme contributes to the performance gain from the baseline.
In addition, as shown in Table~\ref{tab: table6}, we also evaluated the performance of using the three schemes together without fidelity estimation (FE) by replacing the learned fidelity with the pre-defined ground-truth fidelity. We see that using the proposed fidelity values further improves the performance compared with using the static pre-defined fidelity values.

\begin{figure}[t]
	\centering
	\includegraphics[width=1\linewidth]{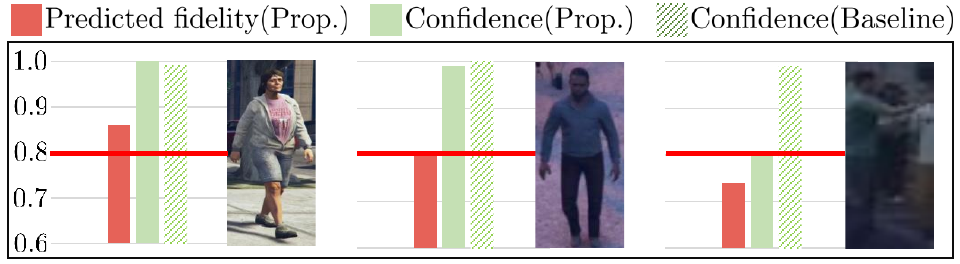}
	\caption{\normalsize{The fidelity and detection confidence estimated by the proposed fidelity adaptive training. The three instances have similar ground-truth fidelity values around 0.8. However, they are assigned different fidelity values according to their actual appearance or visibility. The initial scores of the detection confidence are also changed accordingly.}}
	\label{fig: figure6}
\end{figure}

Figure~\ref{fig: figure6} also demonstrates the effectiveness of the proposed FE by showing the estimated fidelity values and the detection confidence scores. The three examples of person instances have similar BRISQUE scores and hence similar pre-defined fidelity values around 0.8. However, we observe their appearance and visibility are different from one another, which actually affect the performance of person search. For example, the first instance exhibits a relatively clear appearance of person, and yields a much higher learned fidelity value than the ground-truth one by using the proposed FAT. 
On the contrary, the third instance has relatively degraded visibility with blur and low contrast. In such a case, the network assigns a low fidelity value than the ground-truth according to its visibility. The detection confidence score is also forced to be a lower value compared to that of the baseline, by the confidence loss $\mathcal{L}_\mathrm{con}$ in~\eqref{eq: equation2} in FAT.

\begin{table}[t]
	\centering
	\small{
	\begin{tabular}{ccc|cc|cc}
		\toprule
        {\multirow{2}{*}{$\mathcal{L}_{\mathrm{dom}}$}} & {\multirow{2}{*}{\text{DGN}}} & {\multirow{2}{*}{$\mathcal{L}_{\mathrm{sep}}$}} & \multicolumn{2}{c|}{CUHK-SYSU}  & \multicolumn{2}{c}{PRW} \\  \cline{4-7}
            & & & {mAP} & {Top-1} & {mAP} & {Top-1} \\
		\midrule\midrule[.1em]
		{ }          & { }          & { }        & {66.7}        & {71.0}            & {20.9}        & {76.0} \\ 
	{\checkmark} & { }          & { }        & {66.9}        & {70.4}            & {21.1}        & {77.5}\\
		{\checkmark} & {\checkmark} & { }        & {69.0}        & {72.1}            & {23.2}        & {77.0}\\
		{\checkmark} & {\checkmark} & {\checkmark } & \textbf{69.3} & \textbf{72.7}  & \textbf{24.8} & \textbf{79.8}\\
		\bottomrule
	\end{tabular}}
	\caption{\normalsize{Ablation study of $\mathcal{L}_{\mathrm{dom}}$, domain-guided normalization (DGN), and $\mathcal{L}_{\mathrm{sep}}$.}}
\label{tab: table7}
\end{table}

\begin{table}[t!]
	\centering
	\resizebox{0.45\textwidth}{!}{
	\begin{tabular}{c|cc|cc}
		\toprule
		{\multirow{2}{*}{Method}} & \multicolumn{2}{c|}{CUHK-SYSU}  & \multicolumn{2}{c}{PRW} \\ \cline{2-5}
		& {mAP} & {Top-1} & {mAP} & {Top-1}\\
		\midrule\midrule[.1em]
		{Instance Norm. }      & {68.2}        & {70.6}        & {22.0}        & {75.4} \\ 
		{Domain-guided Norm. } & \textbf{69.0} & \textbf{72.1} & \textbf{23.2} & \textbf{77.0} \\
		\bottomrule
	\end{tabular}}
	\caption{\normalsize{The performance of the domain guided normalization compared to the instance normalization.}}
	\label{tab: table8}
\end{table}

\begin{figure}[t]
	\centering
	\subfloat{
		\includegraphics[width=0.084\linewidth]{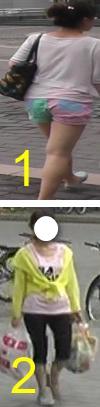}}
	\subfloat{
		\includegraphics[width=0.43\linewidth]{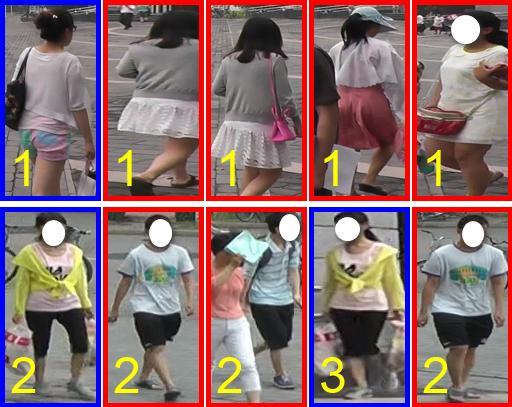}}
  	\subfloat{
		\includegraphics[width=0.43\linewidth]{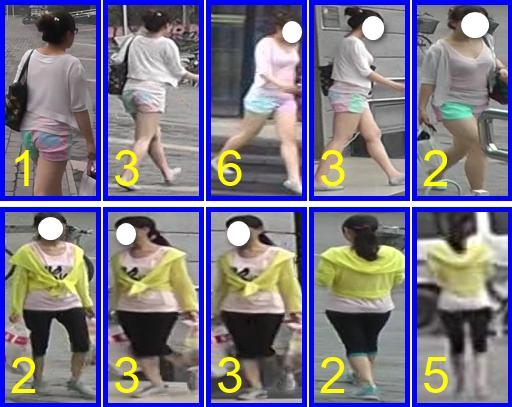}}
	\caption{\normalsize{Comparison of the qualitative performance. Query person images (left) and the Top-5 matching results of the baseline (middle) and the proposed method (right). The true and false matching results are depicted in blue and red, respectively. The camera IDs are indicated in yellow.}}
	\label{fig: figure7}
\end{figure}

\subsubsection{Effect of Domain Invariant Feature Learning.}
Table~\ref{tab: table7} compares the performance by incorporating the domain loss ($\mathcal{L}_{\mathrm{dom}}$), domain-guided normalization (DGN), and domain separation loss ($\mathcal{L}_{\mathrm{sep}}$), respectively. We see that every method improves the performance. 
Note that the instance normalization is widely employed for DG that serves to alleviate the style variations between different domains. We also conducted an experiment to see the effect of DGN compared to the instance normalization. Table~\ref{tab: table8} shows the results where we see that the proposed DGN outperforms the instance normalization on both target datasets. It means that the proposed DGN preserves more useful information for person re-identification while alleviating the information associated with domain variations more effectively.

Note that DIL achieves a relatively high performance gain on the PRW dataset compared to the CUHK-SYSU dataset. Whereas the images in CUHK-SYSU are captured by a single camera within relatively short time durations, the images in PRW are captured by 6 different cameras possibly comprising 6 different domains. Therefore, it becomes more challenging in the PRW dataset to find the person instances having the same ID across different domains. Accordingly, a relatively high impact of DIL is observed in PRW where the domain-related features are suppressed while the ID-specific features are exploited. 
Figure~\ref{fig: figure7} verifies the cross-domain discriminative capability of the proposed method by showing the Top-5 matching results to the query images in PRW dataset. The true and false matching results are depicted in blue and red, respectively, and the camera IDs are indicated in yellow. The baseline method tends to match the person instances from the same camera to the query with high similarity values, and usually fails to find the correct persons across different domains. On the contrary, the proposed method effectively alleviates the camera-dependent information, and therefore, successfully finds the persons across different cameras. 

\section{Conclusion}
In this paper, we introduced a novel framework of domain generalizable person search that uses an automatically labeled unreal dataset only for training to avoid the time-consuming and labor-intensive data labeling and the privacy issues in real datasets. 
To alleviate the domain gaps between the unreal and real datasets, we trained an end-to-end network by estimating the fidelity of person instances simultaneously. 
We also devised the domain-invariant feature learning scheme to encourage the network to suppress the domain-specific information while learning the ID-related features more faithfully. 
Experimental results showed that the proposed method achieves the competitive performance compared to the existing person search methods, even though it is applicable to arbitrary unseen datasets without any prior knowledge of the target domain and additional re-training burdens.

\section{Acknowledgments}
This work was supported by the Institute of Information \& communications Technology Planning \& Evaluation (IITP) grant funded by the Korea government (MSIT) (No.2020-0-01336, Artificial Intelligence Graduate School Program(UNIST)) and (No.2021-0-02068, Artificial Intelligence Innovation Hub).

\bibliography{aaai24}

\end{document}